# Better Predict the Dynamic of Geometry of In-Pit Stockpiles Using Geospatial Data and Polygon Models

Mehala.Balamurali[1], Konstantin M. Seiler[2]

[1,2]Australian Centre for Field Robotics, University of Sydney

Email: mehala.balamurali@sydey.edu.au[1], konstantin.seiler@sydney.edu.au[2]

Modelling stockpile is a key factor of a project economic and operation in mining, because not all the mined ores are not able to mill for many reasons. Further, the financial value of the ore in the stockpile needs to be reflected on the balance sheet. Therefore, automatically tracking the frontiers of the stockpile facilitates the mine scheduling engineers to calculate the tonnage of the ore remaining in the stockpile. This paper suggests how the dynamic of stockpile shape changes caused by dumping and reclaiming operations can be inferred using polygon models. The presented work also demonstrates how the geometry of stockpiles can be inferred in the absence of reclaimed bucket information, in which case the reclaim polygons are established using the diggers GPS positional data at the time of truck loading. This work further compares two polygon models for creating 2D shapes.

**INTRODUCTION**

Detailed and up-to-date knowledge about material contained in stockpiles is crucial for modern mining operations. Knowledge about stockpiles is used for strategic planning, to achieve blend targets as well as for business reporting. Tracking the stockpile contents purely incrementally by counting the number of truck loads that have been dumped and reclaimed quickly leads to drifts caused by tracking errors from the fleet management system as well as variable payloads across individual track loads.

Conventional methods for estimating the volume of stockpiles assume the geometry shape of the stockpiles are regular such as rectangular, triangular prism and trapezoidal (Bajtala et al, 2012) and therefore use mathematical models such as trapezoidal, Simpson-based, cubic spline, and cubic Hermite formulas. These methods use three dimensions points collected using the unmanned aerial vehicle (UAV) and the digital elevation models (DEM) (Uysal, et al 2015; Hamzah et al 2011; Ulvi, 2018). However, the surface of stockpiles is not often a regular shape. Thus, the models using mathematical calculations are prone to failure to infer the volume of stockpiles correctly (Zhao et al., 2012).

UAV based photogrammetric point clouds are sometimes used to estimate the volume of stockpiles with irregular shapes. High-resolution and dense 3D points generated by UAV have advanced 3D surface models and thus provide detailed geometry information of irregular stockpiles (Mancini et al, 2017, Sayab et al, 2018; Raeva et al, 2016, Ref 17; Ref 18). Although 3D models from UAV-based photogrammetry provide high accuracy, measuring stockpile volumes explicitly using photogrammetry technique or any other sensor data that collects 3D point clouds such as LIDAR is time consuming and the methods need high storage and additional data collection techniques. In addition to this, most of the studies regarding the stockpile volume estimation are limited to product stockpiles.

Modern mine automation systems provide telemetry data from mobile equipment which is available close to real-time. This allows for fine-grained inferencing of the movements and actions of vehicles.

This paper investigates the possibility to use high resolution positional data from global navigation satellite systems (GNSS) such as GPS to perform volumetric estimates of a in pit stockpile's content.

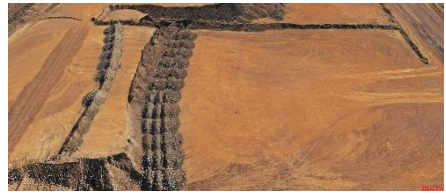

*Figure 1 A typical in-pit stockpile shows that material can be dumped from either the low side (right-hand side) or high side (left-hand side). Reclaiming only happens from the low side.*

Figure 1 shows a typical in-pit stockpile as it is considered in this work. The stockpile consists of a low side and a high side (skyway). Material can be added to the stockpile from the low side through paddock-dumping where truckloads are placed on the ground in individual mounds. Alternatively, material is added from the skyway by dumping it over the edge, potentially further building up an area which already contains material from paddock-dumping. When reclaiming, material is always recovered from the low side of the stockpile over the entire height of the available material.

By using information from a site's fleet management system (FMS) and position data from vehicles, the location of a dump event can be inferred accurately. Similarly, by combining position information from wheel loaders and FMS data, the position material is reclaimed from can be inferred. This work combines the information about individual load and dump events to infer the shape of the stockpile throughout the operation.

The main objective of this study is to create timely 2D geometric shapes of stockpiles using truck dump and load events at stockpiles. A Convex-hull method (Jarvis, 1973, 1977) was used to create the polygons around dump points that demonstrate how the different parcels of ore grades that are added to and reclaimed from the stockpile over different time periods. Convex-hulls are widely used to create d-dimension convex polygons that encloses all the given points (Dunder et al, 2008; Zhongliang and Yuefeng, 2012; Jingfan Fan et al, 2013; Hazra et al, 2004). These polygons were tested on an active in-pit stockpile at Pilbara region in Western Australia, where the material was dumped and reclaimed simultaneously. The dump and reclaiming polygons were created using truck GPS positions and loader bucket reclaim positions respectively.

As comparison, a second polygon model based on alpha shapes was created to represent the stockpile geometry. Alpha-shape polygons are derived from the Delaunay triangulation for a given set of points (Edelsbrunner, 1983; Edelsbrunner et al, 1994). The polygon shape is controlled by a parameter that defines how tightly the boundary hugs around points (Edelsbrunner et al., 2006, Weiqiang and Hong, 2010).

Further, this study demonstrates how the geometry of stockpiles can be inferred in absence of loader bucket information, in which case the reclaim polygons are established purely using the digger GPS positional data at the time of truck loading. Aerial photographs were used to validate the results at different time points.

## METHODOLOGY

This section describes the data used in this study and the techniques of Convex hull and Alpha shape polygons which are central to this work. Publicly available polygon methods were tested. Two algorithms are provided in this study. The Algorithm 1 describes how the 2D geometry of the stockpile grows continuously with the continuous dumps only or how the geometry of the stockpile continuously shrinks with continuous reclaims only. Algorithm 2 describes the dynamic of 2D geometry of stockpile by simultaneously considering dump events followed by the reclaim events or other way.

**Data**

Data was collected at a stockpile in an iron ore surface mine in Australia's Pilbara region from March to December 2019. Truck GPS positions indicating dump events were collected using timestamps provided by the fleet management system and by selecting samples where the truck is standing still. Similarly, bucket reclaim positions and the digger GPS positions were collected at the time of a truck being present for loading at the stockpile. These raw positional data points were then used as input to create dump and load polygons for a given time period. Different intervals were used to create the polygons.

**Convex hull**

Convex hulls are widely used in many fields to create d-dimension convex polygons that enclose all the given points. The main idea behind a convex hull comes from the definition of a convex polygon. A convex polygon is a polygon with all its interior angles less than 180°. By using this definition, every regular polygon is convex. If one angle has more than 180 degrees, the polygon is considered to be concave. A convex hull uses the same principle as convex polygon applied to set of points. For instance, a convex hull is the smallest convex polygon containing all the points of a set. While simple and well understood, convex hulls come with the limitation that they are unable to represent concave shapes.

**Alpha shape**

Unlike convex hull, alpha shape gives a better precision for finding accurate outline of a set of points. Alpha-polygon is a simple polygon in which all internal angles are less than 180+alpha. When alpha is equal to zero, then the polygon is convex hull polygon. Alpha shape polygons are derived from the Delaunay triangulation for a given set of points.

In mathematics and computational geometry, a Delaunay triangulation for a given set P of discrete points in a plane is a triangulation DT(P) such that no point in P is inside the circumcircle of any triangle in DT(P). Delaunay triangulations maximize the minimum angle of all the angles of the triangles in the triangulation. Computing the Delaunay triangulation of a point set consists in linking all the points in a way of creating triangles linking all the points then removing all the triangles for which at least one edge exceeds alpha in length.

**Dump or reclaim polygon**

The convex hull or alpha shapes likely break (or overestimate reclaiming or dumping) if the dump and reclaim actions don't occur in straight line patterns. As can be seen in Figure 2 while some trucks dumped at high side (Red) and some dumped at lower side (green points) at the same time interval or there is any kind of digging that's not a single contiguous "chunk".

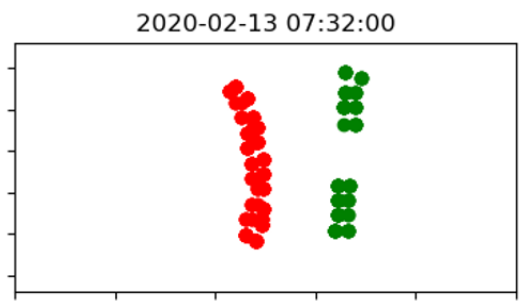

*Figure 2: Dumping happens as different clusters; red points show dumps at high end of stockpile and green points show the dumps at lower side of stockpile.*

To provide 2D polygons for spatially separated chunks of dump and reclaim events, dump and reclaim positions within a time range were clustered using DBSCAN. DBSCAN is a density-based spatial clustering method commonly used for GPS trajectory data processing (Ester, 1996). It is based on the observation that points surrounding a dump or dig event usually form a high-density cluster. Hereby two parameters are used to define density: the search distance (Eps) and the minimum number of points (MinPts) within an area defined by the distance threshold. Points which satisfy those conditions related to density were grouped to form dump and reclaim clusters. Then the polygon model is then applied to each cluster.

The proposed Algorithm 1 for dump or reclaim polygons describes how the polygon grows during continuous dumping and shrinks during continuous reclaiming. This algorithm uses Scikit learn implementation for DBSCAN and publicly available implementation for convex hull or alpha shape polygons (Sean and Kevin).

| ALGORITHM 1: DUMP/RECLAIM POLYGON |
|---|
| Step 1: Define variables $t_0$, $\Delta t$, $t_f$ and $T_s$ for start of event interval, event duration, end of event interval and timestamp for stopping computation; Array 'Cluster' for cluster labels and array GPS for truck GPS positions or bucket reclaim positions (x, y, timestamp); DBSCAN, Polygon and Plot-hull are functions. Here the function Polygon can be replaced with Convex hull or Alpha shape implementation. Plot-hull is to plot the polygons. |
| Step 2: Start with initial values $t_0$, $\Delta t$ |
| Step 3: Repeat steps below until $t_f = T_s$ |
|     3.1    $t_f = t_0 + \Delta t$ |
|     3.2    Locations = GPS [timestamp] $\subseteq (t_0, t_f)$ |
|     3.3    Cluster = DBSCAN (Locations) |
|     3.4    For c = 1: length (unique (Cluster)) <br>         hull = Polygon (Locations [c]) <br>         plot-hull |
|     3.5    $t_0 = t_f$ |

The Algorithm 2 is proposed to see the dynamics of a stockpile's 2D shape by tracing both regions of dumps and reclaims when the reclaim polygons simultaneously remove the dumps. The algorithm finds the area of dump polygon intersects with the reclaim polygons and then removes the dumps cover by the corresponding reclaim polygon. This algorithm also uses diggers' GPS position at the time of truck loading to create 2D geometry of stockpile shape in the absence of reclaimed bucket position data.

| ALGORITHM 2: |
|---|
| Step 1: Define variables $t_0$, $\Delta t$, $t_f$ and $T_s$ for start of event interval, event duration, end of event interval and timestamp for stopping computation; Array 'Cluster' for cluster labels and array 'GPS_TRUCK' for truck GPS positions/ dump locations (x, y), 'BUCKET' for bucket dig positions and 'GPS_DIGGER' for diggers GPS position with timestamps; DBSCAN, Polygon are functions. |
| Step 2: Start with initial values $t_0$, $\Delta t$ |
| Step 3: Repeat steps below until $t_f = T_s$ |
|     3.1    $t_f = t_0 + \Delta t$ |
|     3.2    DL = GPS_TRUCK $\subseteq (t_0, t_f)$ |
|     3.3    BL = BUCKET $\subseteq (t_0, t_f)$ |
|     3.4    WL_GPS = GPS_DIGGER $\subseteq (t_0, t_f)$ |
|     3.5    If number of rows in (BL & WL_GPS) = 0 |
|         3.5.1    Cluster = DBSCAN (DL) |
|         3.5.2    For c=1: length (unique (Cluster)) <br>             D_hull = Polygon (DL[c]) |

> 3.6 Elseif number of rows in BL > 0
>   3.6.1 R_hull = Polygon (BL)
>   3.6.2 inhull_samples= D_hull ∩ R_hull
>   3.6.3 remaining_DL=DL- inhull_samples
>   3.6.4 Cluster=DBSCAN (remaining_DL)
>   3.6 .5 For c= 1: length (unique (Cluster))
>         D_hull = Polygon (remaining_DL[c])
>
> 3.7 Elseif number of rows in BL= 0 & number of rows in WL_GPS > 0
>   3.7.1 R_hull = Polygon (WL_GPS)
>   3.7.2 inhull_samples= D_hull ∩ R_hull
>   3.7.3 remaining_DL= DL- inhull_samples
>   3.7.4 Cluster=DBSCAN (remaining_DL)
>   3.7.5 For c= 1 : length (unique (Cluster))
>         D_hull = Polygon (remaining_DL[c])
> 3.8 $t_0 = t_f$

## RESULTS AND DISCUSSION

**Dump and reclaim polygons using Algorithm 1**

Figure 3 shows how the different parcels of dumps are clustered using DBSCAN, thus creating different polygons around the dump.

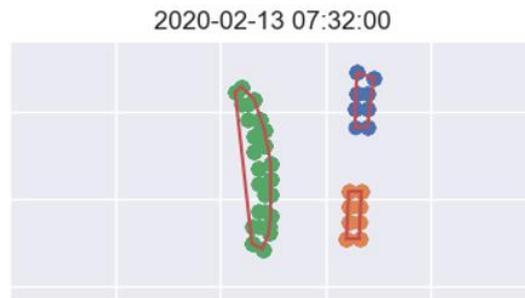

*Figure 3 shows the convex hull polygons on clustered dump positions*

The polygons in Figure 4a show how the stockpile area is continuously growing as the trucks are continuously dumping material at the given stockpile. Similarly, the polygons in Figure 4b show how the reclaim area is growing continuously with the continuous bucket reclaim points from the same stockpile. Each polygon represents the results obtained by Algorithm 1 that used convex hull polygon model with the truck dump points or bucket reclaim points for every two hours interval and 30 minutes intervals respectively. Colours of the polygons represent the oldest to latest polygons in both figures. Therefore, the latest dump polygon in Figure 4a (colour number 14) shows the stockpile area with full of material and the latest reclaim polygon in Figure 4b shows the empty stockpile area. The dump positions were inferred from the truck GPS positions.

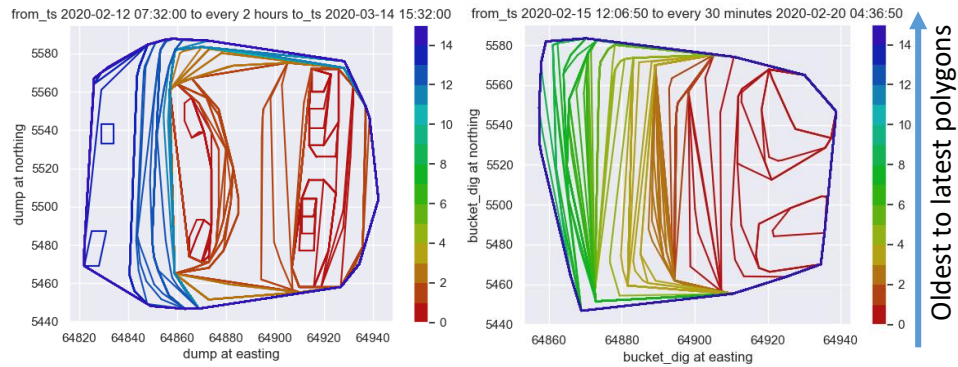

*Figure 4 (Left)Convex hull dump polygons around the dump points for every two hours interval. (Right) reclaim polygons around the reclaim positions for every 30 minutes. Number on colour bar 0 to 14 shows oldest to latest polygons.*

Computing the convex hull of sets of dumps or reclaim points is a way to represent the region occupied by these points. The convex hull of set of points is convex polygon with the minimum area that includes all these points.

**Dynamic of stockpile shape changes caused by dumping and reclaiming operations**

Figure 5 shows the results obtained by Algorithm 2. In this example, the proposed algorithm used the convex hull polygon. The intersecting area of the dump polygons (sky-blue and grey) shows the area where the dumps at high side (green) had been overlaid on earlier dumps (pink) at low side. The plots in Figure 5 show how the dumps had been continuously removed using the reclaim polygons (red/blue) from the right side to left side at the low side of the stockpile. The plots were produced for every 24 hours period. Reclaim polygons (red) that used bucket reclaim positions, were used to remove the previous dumps. The blue reclaim polygons in Figure 5 were created using digger's GPS positions and the polygon is presented here for a comparison purpose. The comparison between the blue and red polygon shows that the polygons around the digger's GPS positions closely follow the reclaim polygons of bucket reclaim points. Thus, the individual plots in Figure 5 clearly shows that the digger's GPS positions can be used as an alternative to bucket reclaim positions to create reclaim polygons with the correct off set values. This alternative data is beneficial for removing the dumps during loading because sometimes diggers frequently go into off-nominal states due to disturbances or faults and thus fail to record all bucket reclaim positions. The digger used in this study was a wheel loader.

It can be seen in Figure 5, even though the dumps had been reclaimed, still there were some false dump points at the lower side of the stockpile and therefore the plots still show the false dump polygon at the low side of the stockpile instead of only showing the dump polygon (sky-blue) at high side. The model results can be improved by identifying the exact dump and reclaim positions using the calibrated truck and bucket positions and removing outliers. In addition to this, using improved polygon models with small time intervals may help define the dump and reclaim regions accurately and therefore can remove the dumps correctly. Optimising the parameters can further help to improve the results because the convex hull shape and DBSCAN depends on radius parameters and distance thresholds. These are subject to further investigation. The aerial image given in Figure 5 was captured in December at the same stockpile used in this example. Even though the exact time this figure was taken is not known, the remaining stockpile seen in the high side area supports the results, sky-blue polygon area, obtained by the model in December.

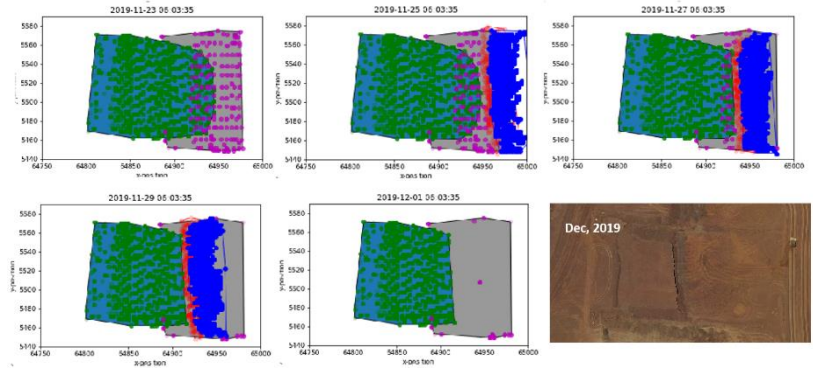

*Figure 5: Green and pink points represent the dump (truck GPS position) locations and sky-blue and grey polygons represent the corresponding dump polygons at high and low side of the stockpile respectively. Red and blue points show the reclaim locations by bucket dig locations and diggers GPS positions and the corresponding colour polygons show the reclaimed area.*

**Tracking the dynamic of 2D geometry of stockpile in the absence of bucket reclaim positions**

The plots in the left and right columns of Figure 6 compare how the stockpile geometry shape can be inferred in the absence of bucket reclaim positions. It has been proved from the trace of diggers GPS positions (blue points) in Figure 6 that the diggers continuously had loaded material in that area, but the bucket reclaim positions had not been recorded (no red points). Thus, when Algorithm 2 was set to only use bucket reclaim positions, the model left the dumps unchanged (Figure 6 left column). On 2019-06-01 the plot on the left column shows some new dumps underneath the existing high side dumps. This is not possible and thus the results again prove that the diggers GPS positions shown in Figure 6 are associated with some reclaiming action. The plots in the right column of Figure 6 show how the reclaim polygon that used diggers GPS positions, had removed the dumps continuously. As mentioned before, even though we could still see the presence of dump polygon at the low side due to the falsely left over dump points, the results reasonably justify the dumps that appeared on 2019-06-01 were the new dumps that were dumped at the low side of the stockpile next to the edge of the high side (right plots in Figure 6).

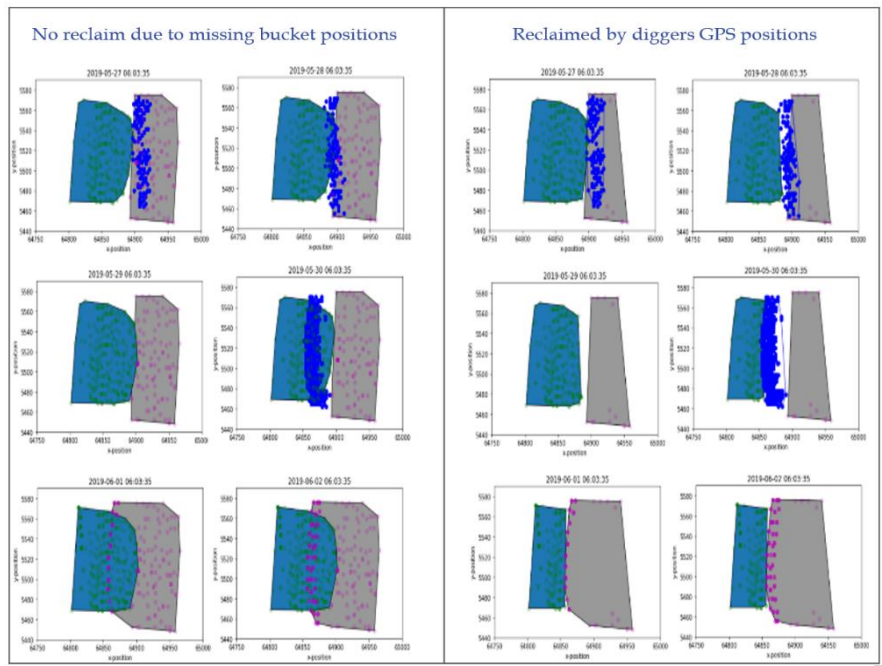

*Figure 6: Green and pink points represent the dump (truck GPS position) locations and sky-blue and grey polygons represent the corresponding dump polygons at high and lower side of the stockpiles respectively. Blue*

*points in column left show the diggers GPS position and blue polygons in right column shows the reclaim polygon using diggers GPS*

The aerial image (Fig 7) taken in June at the same stockpile supports the model results provided in Figure 6. Therefore, a conclusion can be drawn that diggers GPS positions can be used to create reclaim polygons in the absence of bucket reclaim positions.

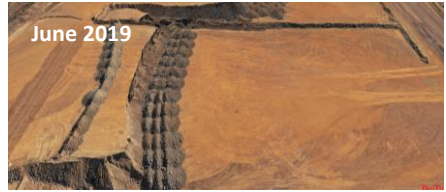

*Figure 7 shows the aerial image of stockpile where new dumps are happening at the lower side of the stock pile.*

**Alpha shape vs Convex hull**

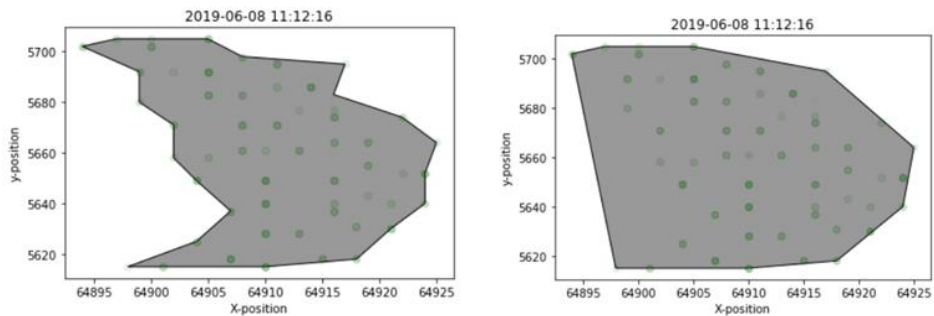

*Figure 8 compares the Alpha shape (left) and Convex hull (right) polygons around the dump points (green)*

The alpha-shapes methodology presented here represents an additional tool for quantifying 2D shape complexity across dumps characterised by high spatial disparity. As can be seen in Figure 8 the boundaries of the truck GPS positions (dump positions) are better represented by the alpha shape polygon compared to the convex hull polygon. The convex hull (Fig. 8b) occupies a space equal to or larger than that of the underlying dump locations. In contrast, very fine alpha shape (Fig. 8a) defined a 2D space smaller than the area defined by convex hull. This was obtained by some 'optimal' level of refinement. Figure 9 shows the alpha shape polygon around the dumps on satellite image.

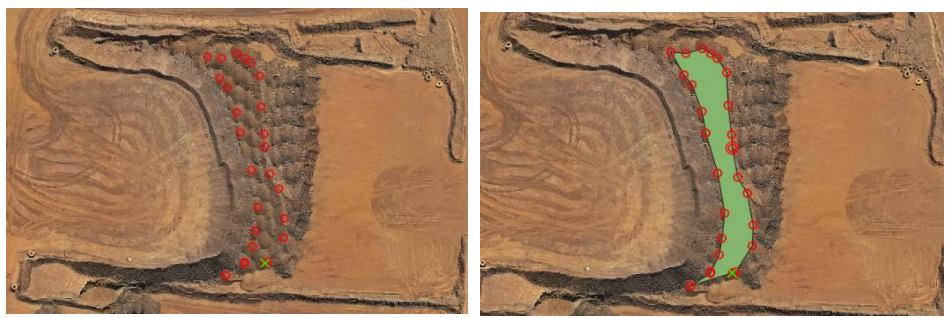

*Figure 9: Alpha shape polygon around dumps are shown in satellite imagery*

# CONCLUSION

The pilot work presented here is an important addition to the geologist's tool kit, providing a guidance of how the stockpile grows and ablates with time. The model suggests that the calibrated diggers GPS positions can be used to create reclaim polygons when the buckets' information are missing and therefore the model can help to track the 2D geometry of the stockpile. Compared to convex hull, alpha shape proves to be especially useful in representing the topographical complexity of a stockpile surface. Creating the dump and reclaim polygons in the short time interval will help to understand the dynamic of 2D geometry better. Having high frequency aerial images can be a great tool for validating the model results. Further investigation is needed to improve the model outputs.

# ACKOWLEDGMENT

This work has been supported by the Australian Centre for Field Robotics and the Rio Tinto Centre for Mine Automation, the University of Sydney.